\documentclass[11pt,a4paper]{article}
\usepackage[hyperref]{acl2020}
\usepackage{times}
\usepackage{latexsym}

\usepackage{microtype}
\usepackage{adjustbox}
\usepackage{makecell}
\usepackage{multirow}
\usepackage{placeins}
\usepackage{amsmath,amssymb,amsthm, amsfonts,color}
\aclfinalcopy 
\def\aclpaperid{34} 

\title{Comparative Analysis of Text Classification Approaches in Electronic Health Records}

\author{ Aurelie Mascio\thanks{\hspace{1mm} These two authors contributed equally.}
         \hspace{3mm} Zeljko Kraljevic\footnotemark[1]\\
         \hspace{3mm}\textbf{Daniel Bean}  \hspace{3mm}\textbf{Richard Dobson} \hspace{3mm}\textbf{Robert Stewart} 
         \hspace{3mm}\textbf{Rebecca Bendayan} \hspace{3mm}\textbf{Angus Roberts} \vspace{3mm}\\ 
   Department of Biostatistics \\
  \& Health Informatics \\
  King's College London, UK \vspace{3mm}\\
  \href{mailto:aurelie.mascio@kcl.ac.uk}{aurelie.mascio@kcl.ac.uk} \hspace{3mm} \href{mailto:zeljko.kraljevic@kcl.ac.uk}{zeljko.kraljevic@kcl.ac.uk}
  \\}

\date{}

\begin{document}
\maketitle
\begin{abstract}
    Text classification tasks which aim at harvesting and/or organizing information from electronic health records are pivotal to support clinical and translational research. However these present specific challenges compared to other classification tasks, notably due to the particular nature of the medical lexicon and language used in clinical records.\\
    Recent advances in embedding methods have shown promising results for several clinical tasks, yet there is no exhaustive comparison of such approaches with other commonly used word representations and classification models.\\
    In this work, we analyse the impact of various word representations, text pre-processing and classification algorithms on the performance of four different text classification tasks. The results show that traditional approaches, when tailored to the specific language and structure of the text inherent to the classification task, can achieve or exceed the performance of more recent ones based on contextual embeddings such as BERT.
\end{abstract}

\vspace{5mm}
\section{Introduction}
Clinical text classification is an important task in natural language processing (NLP) \citep{yao_clinical_2019}, where it is critical to harvest data from electronic health records (EHRs) and facilitate its use for decision support and translational research. Thus, it is increasingly used to retrieve and organize information from the unstructured portions of EHRs \citep{mujtaba_clinical_2019}.\\ Examples  include tasks such as: (1) detection of smoking status \citep{uzuner_identifying_2008}; (2) classification of medical concept mentions into family versus patient related \citep{dai_family_2019}; (3) obesity classification from free text \citep{uzuner_recognizing_2009}; (4) identification of patients for clinical trials \citep{meystre_automatic_2019}.\\
Most of these tasks involve mapping mentions in narrative texts (e.g. “pneumonia”) to their corresponding medical concepts (and concept ID) generally using the Unified Medical Language System (UMLS) \citep{bodenreider_unified_2004}, and then training a classifier to identify these correctly (e.g. “pneumonia positive” versus “pneumonia negative”) \citep{yao_clinical_2019}.

Text classification performed on medical records presents specific challenges compared to the general domain (such as newspaper texts), including dataset imbalance, misspellings, abbreviations or semantic ambiguity \citep{mujtaba_clinical_2019}.\\
Despite recent advances in NLP, including neural-network based word representations such as BERT \citep{devlin_bert_2019}, few approaches have been extensively tested in the medical domain and rule-based algorithms remain prevalent \citep{koleck_natural_2019}. Furthermore, there is no consensus on which word representation is best suited to specific downstream classification tasks \citep{si_enhancing_2019, wang_clinical_2018}.

The purpose of this study is to analyse the impact of numerous word representation methods (bag-of-word versus traditional and contextual word embeddings) as well as classification approaches (deep learning versus traditional machine learning methods) on the performance of four different text classification tasks. To our knowledge this is the first paper to test a comprehensive range of word representation, text pre-processing and classification methods combinations on several medical text tasks.

\pagebreak
\section{Materials \& Methods}
\subsection{Datasets and text classification tasks}
\label{sec:tasks}
    In order to conduct our analysis we derived text classification tasks from MIMIC-III (Multiparameter Intelligent Monitoring in Intensive Care) \citep{johnson_mimic-iii_2016},  and the Shared Annotated Resources (ShARe)/CLEF dataset \citep{mowery_task_2014}. These datasets are commonly used for challenges in medical text mining and act as benchmarks for evaluating machine learning models \citep{purushotham_benchmarking_2018}.
    
    \paragraph{MIMIC-III dataset}
    MIMIC-III \citep{johnson_mimic-iii_2016} is an openly available dataset developed by the MIT Lab for Computational Physiology. It comprises clinical notes, demographics, vital signs, laboratory tests and other data associated with 40,000 critical care patients.\vspace{5mm}

    We used MedCAT \citep{kraljevic_medcat_2019} to prepare the dataset and annotate a sample of clinical notes from MIMIC-III with UMLS concepts \citep{bodenreider_unified_2004}.
    We selected the concepts with the UMLS semantic type Disease or Syndrome (corresponding to T047), out of which we picked the 100 most frequent Concept Unique Identifier (CUIs, allowing to group mentions with the same meaning). For each concept we then randomly sampled 4 documents containing a mention of each concept,  resulting in 400 documents with 2367 annotations in totals. The 100 most frequent concepts in these documents were manually annotated (and manually corrected in case of disagreement) for two text classification tasks:
    \begin{itemize}
        \item Status (affirmed/other, indicating if the disease is affirmed or negated/hypothetical); 
        \item Temporality (current/other, indicating if the disease is current or past).
    \end{itemize}
    Such contextual properties are often critical in the medical domain in order to extract valuable information, as evidenced by the popularity of algorithms like ConText or NegEx \citep{harkema_context_2009,chapman_simple_2001}.\vspace{5mm}
    
    Annotations were performed by two annotators, achieving an overall inter-annotator agreement above 90\%. These annotations will be made publicly available.
    
    \paragraph{ShARe/CLEF (MIMIC-II) dataset}
    The ShARe/CLEF annotated dataset proposed by \citet{mowery_task_2014} is based on 433 clinical records from the MIMIC-II database \citep{saeed_mimic_2002}.
    It was generated for community distribution as part of the Shared Annotated Resources (ShARe) project \citep{elhadad_share_2013}, and contains annotations including disorder mention spans, with several contextual attributes.\\
    For our analysis we derived two tasks from this dataset, focusing on two attributes, comprising 8075 annotations for each:
    \begin{itemize}
        \item Negation (yes/no, indicating if the disorder is negated or affirmed);
        \item Uncertainty (yes/no, indicating if the disorder is hypothetical or affirmed). 
    \end{itemize}
    
    \paragraph{Text classification tasks}
    For both annotated datasets, we extracted from each document the portions of text containing a mention of the concepts of interest, keeping 15 words on each side of the mention (including line breaks).
    Each task is then made up of sequences comprising around 31 words, centered on the mention of interest, with its corresponding meta-annotation (status, temporality, negation, uncertainty), making up four text classification tasks, denoted: 
    \begin{itemize}
        \item MIMIC \textbar \ Status; 
        \item MIMIC \textbar \ Temporality;
        \item ShARe \textbar \ Negation; 
        \item ShARe \textbar \ Uncertainty.
    \end{itemize}   
    
    Table~\ref{tab:classes} summarizes the class distribution for each task.
        \begin{table}[!h]
        \centering
        \begin{adjustbox}{width=.5\textwidth,center}
        \begin{tabular}{cccc}
        \hline
        \textbf{Task} & \textbf{Class 1} & \textbf{Class 2} & \textbf{Total}\\
        \hline
        \makecell{\textbf{MIMIC \textbar \ Status} \\ (1: affirmed, 2: other)} & 1586 (67\%) & 781 (33\%) & 2367\\ \hline
        \makecell{\textbf{MIMIC \textbar \ Temporality} \\ (1: current, 2: other)} & 2026 (86\%) & 341 (14\%) & 2367\\ \hline
        \makecell{\textbf{ShARe \textbar \ Negation} \\ (1: yes, 2: no)} & 1470 (18\%) & 6605 (82\%) & 8075\\ \hline
        \makecell{\textbf{ShARe \textbar \ Uncertainty} \\ (1: yes, 2: no)} & 729 (9\%) & 7346 (91\%) & 8075\\ \hline
        
        \end{tabular}
        \end{adjustbox}
        \caption{Class distribution}
        \label{tab:classes}
        \end{table}

\subsection{Evaluation steps and main workflow}
        \begin{figure*}[!h]
          \centering
          \includegraphics[width=13.5cm]{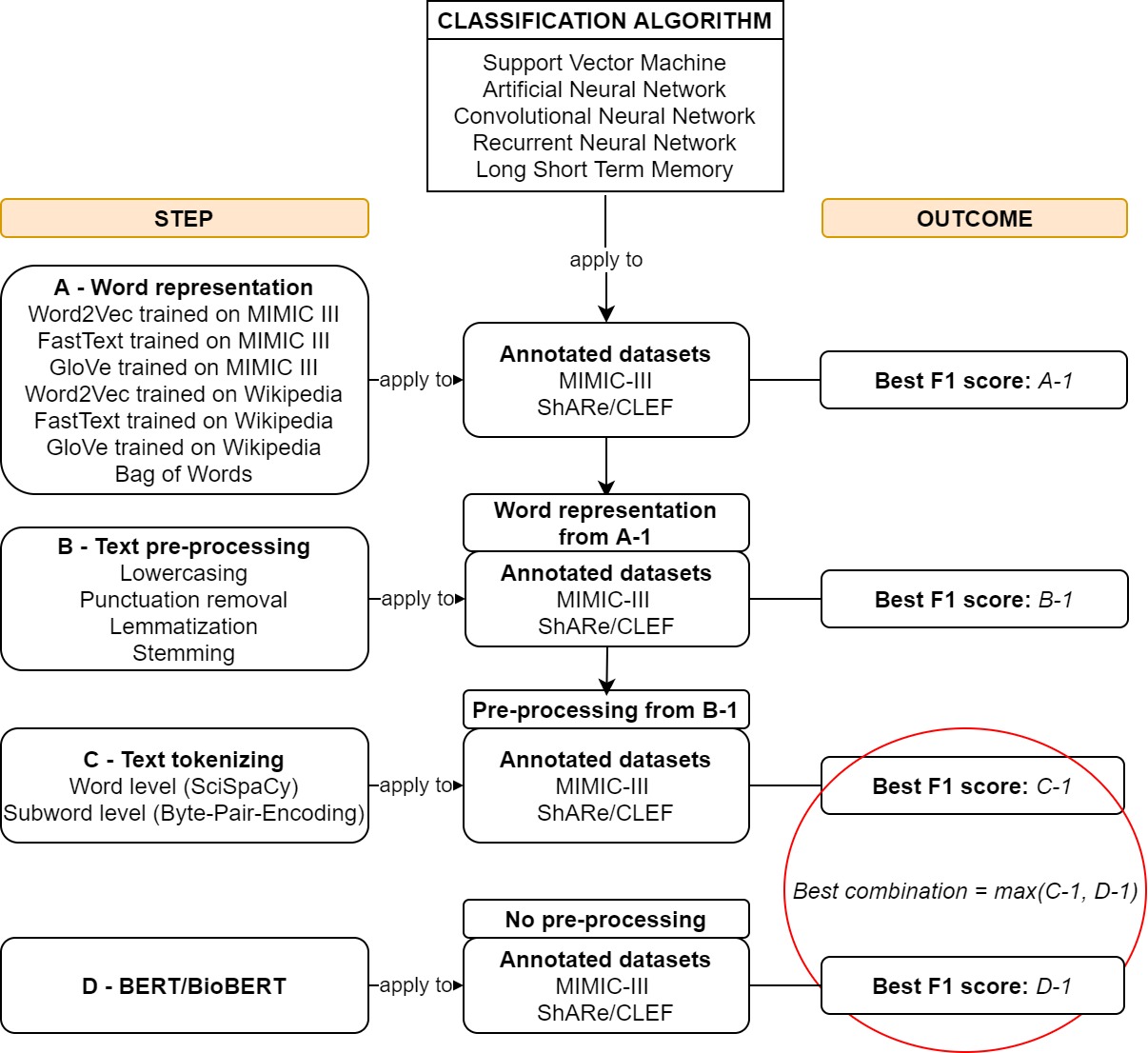}
          \caption{Main workflow}
          \label{fig:workflow}
        \end{figure*}
    We used the four different text classification tasks described in Section~\ref{sec:tasks} in order to explore various combinations of word representation models (see Section~\ref{sec:embeddings}), text pre-processing and tokenizing variations (Section~\ref{sec:preprocessing}) and classification algorithms (Section~\ref{sec:algos}).
    In order to evaluate the different approaches we followed the steps detailed in Table~\ref{tab:eval} and Figure~\ref{fig:workflow} for all four classification tasks.
    
    \begin{table}[!h]
        \centering
        \begin{adjustbox}{width=.5\textwidth,center}
        \begin{tabular}{ccc}
        \hline
        \textbf{Step} & \textbf{Description} & \makecell{\textbf{Outcome} \\ (best F1)}\\
        \hline
        A &
        \makecell{Run all bag-of-word and traditional embeddings\\ + classification algorithms and select the \\
                best combination (using baseline methods for \\ text pre-processing and tokenization)}
        & A-1\\ \hline
        B &
        \makecell{Using A-1 as the new baseline model, test\\ different pre-processing methods (lowercasing, \\punctuation removal, lemmatization, stemming)} &
        B-1\\ \hline
        C &
        \makecell{Using B-1 as the new baseline model, compare various \\ tokenizers (word and subword level)} &
        C-1\\ \hline
        D &
        \makecell{Test contextual embedding approaches: \\ BERT (base, uncased) and BioBERT} &
        D-1\\
        \hline
        \end{tabular}
        \end{adjustbox}
        \caption{Evaluation steps}
        \label{tab:eval}
    \end{table}
    
    For each step we measured the impact by evaluating the best possible combination, based on the average F1 score (weighted average score derived from 10-fold cross validation results).

\subsection{Word representation models}
\label{sec:embeddings}
    Word embeddings as opposed to bag-of-words (BoW) present the advantage of capturing semantic and syntactic meaning by representing words as real valued vectors in a dimensional space (vectors that are close in that space will represent similar words). Contextual embeddings go one step further by capturing the context surrounding the word, whilst traditional embeddings assign a single representation to a given word.
    
    For our analysis we considered four off-the-shelf embedding models, pre-trained on public domain data, and compared them to the same embedding models trained on biomedical corpora, as well as a BoW  representation.\vspace{5mm}
    
    For the traditional embeddings we chose three commonly used algorithms, namely Word2Vec \citep{mikolov_efficient_2013}, GloVe \citep{pennington_glove:_2014} and FastText \citep{bojanowski_enriching_2017}.\\
    We used publicly available models pre-trained on Wikipedia and Google News for all three \citep{yamada_wikipedia2vec_2018}.\\
    To obtain medical specific models we trained all three on MIMIC-III clinical notes (covering 53,423 intensive care unit stays, including those used in the classification tasks) \citep{johnson_mimic-iii_2016}. The following hyperparameters, aligned to off-the-shelf pre-trained models, were used: dimension of 300, window size of 10, minimum word count of 5, uncased, punctuation removed.
    
    For the contextual embeddings we used BERT base \citep{devlin_bert_2019}, and BioBERT \citep{lee_biobert_2019} which are pre-trained respectively on general domain corpora and biomedical literature (PubMed abstracts and PMC articles).
    
    Finally we used a BoW representation as a baseline approach.

\subsection{Text pre-processing and tokenizers}
\label{sec:preprocessing}
    In addition to pre-training several embedding models, we tested two different text tokenization methods, using the following types of tokenizers: (1) SciSpaCy \citep{neumann_scispacy_2019}, a traditional tokenizer based on word detection; and (2) byte-pair-encoding (BPE) adapted to word segmentation that works on subword level \citep{gage_new_1994, sennrich_neural_2016}.
    
    For the word level tokenizer we chose SciSpaCy as it is specifically aimed at biomedical and scientific text processing. We further tested additional text pre-processing: lowercasing, punctuation removal, stopwords removal, stemming and lemmatization.
    
    For the subword BPE tokenizer we used byte level byte-pair-encoding (BBPE) \citep{wang_neural_2019, wolf_huggingfaces_2020}. In this case the only pre-processing performed is lowercasing, whilst everything else including line breaks and spaces is left as is. This approach allows to limit the vocabulary size and is especially useful in the medical domain where a large number of words are very rare. We limited the number of words to 30522, a standard vocabulary size also used in BERT \citep{devlin_bert_2019}. 
    
\subsection{Text classification algorithms} 
\label{sec:algos}
    On all four classification tasks, we tested various machine learning algorithms which are widely used for clinical data mining tasks and achieve state-of-the-art performance \citep{yao_clinical_2019}, namely artificial neural network (ANN), convolutional neural network (CNN), recurrent neural network (RNN), bi-directional long short term memory (Bi-LSTM), and BERT \citep{devlin_bert_2019,wolf_huggingfaces_2020}.\\
    We compared these with a statistics-based approach as a baseline, using a Support Vector Machine (SVM) classifier, a popular method used for classification tasks \citep{cortes_support-vector_1995}.\vspace{5mm}
    
    For Bi-LSTM and RNN, we tested both a standard approach and one that is configured to simulate attention on the medical entity of interest. This custom approach consisted in taking the representation of the network at the position of the entity of interest, which in most cases corresponds to the center for each sequence. We refer to this latter approach as custom Bi-LSTM and custom RNN.
    
    For ANN and statistics-based models, which are limited by the size of the dataset and embeddings (300 dimensions x 31 words x 2300 or 8000 sequences), we chose to represent sequences by averaging the embeddings of the words composing each sequence. This representation method is commonly used and has proven efficient for various NLP applications \citep{kenter_siamese_2016}.
    
    Furthermore, each of these models was tested using different sets of parameters (e.g. varying the support function, dropout, optimizer, as reported in Table~\ref{tab:params}), the ones producing the best performance were selected for further testing and are summarized in Table~\ref{tab:params}.
    \begin{table}[!h]
    \centering
    \begin{adjustbox}{width=.5\textwidth,center}
    \begin{tabular}{ccccc}
        \hline
         & \textbf{SVM} & \textbf{ANN} & \textbf{CNN} & \makecell{\textbf{RNN}\\ \textbf{Bi-LSTM}}\\
        \hline
        \makecell{\textbf{Kernel or} \\\textbf{activation} \\\textbf{function}} & \makecell{ \textbf{Radial basis} \\ Linear \\ Poly \\ Sigmoid} & \makecell{ReLU + \\sigmoid} & \makecell{ReLU \\(with \\max \\pooling)} & N/A\\ \hline
        \textbf{Layers} & N/A & 2 & 3 & 2\\ \hline
        \textbf{Filters} & N/A & N/A & 128 & N/A \\ \hline
        \makecell{\textbf{Hidden units}\\ \textbf{dimensions}} & N/A & 100 & N/A & 300\\ \hline
        \textbf{Dropout} & N/A & \makecell{\textbf{0.5} \\ 0} & \makecell{\textbf{0.5} \\ 0} & \makecell{\textbf{0.5} \\ 0}\\ \hline
        \textbf{Optimizer} & N/A & \makecell{\textbf{Adam} \\ Stochastic \\Gradient \\Descent} & \makecell{\textbf{Adam} \\ Stochastic \\Gradient \\Descent} & \makecell{\textbf{Adam} \\ Stochastic \\Gradient \\Descent}\\ \hline
        \textbf{Learning rate} & N/A & 0.001 & 0.001 & 0.001\\ \hline
        \textbf{Epochs} & \makecell{tolerance:\\ 0.001} & 5000 & 200 & 50\\ \hline
    \end{tabular}
    \end{adjustbox}
    \caption{Classifiers and corresponding parameters evaluated. Parameters highlighted in \textbf{bold} were the ones selected based on performance.}
    \label{tab:params}
    \end{table}

\pagebreak
\section{Results}
\subsection{ Performance comparison for all embedding and algorithm combinations (steps A \& D)}
    In this section we compare the performance of the different embeddings and classification approaches.
    We report the weighted average F1/precision/recall (weighted average value obtained from the 10-fold cross-validation results) for selected combinations on the four classification tasks in Tables~\ref{tab:embeddings} and ~\ref{tab:algos} (full results in Appendix~\ref{sec:appendix1}).\\
    For all word embedding methods tested (Word2Vec, GloVe, FastText), the ones trained on biomedical data show the best performance (see Table~\ref{tab:embeddings}).\vspace{5mm}
    
    For classification algorithms, the best performance is obtained when using the custom Bi-LSTM model configured to target the biomedical concept of interest (see Table~\ref{tab:algos}). Both contextual embeddings (BERT and BioBERT), whether trained on biomedical or general corpora, outperform any other combination of embedding/classification algorithm tested, and give results very close to the customized Bi-LSTM, as shown in Table~\ref{tab:algos}.\vspace{5mm}
    
    This indicates that for tasks incorporating information about the position of the entity of interest in the text (e.g. ShARe which reports disorder mentions span offsets), the custom Bi-LSTM approach performs better than BioBERT, without necessitating any text pre-processing.\\
    On the other hand, when looking at pure text classification, BioBERT shows better performance than a Bi-LSTM approach, and consequently may be preferred for tasks where the sequence of interest is not easily centered on a specific entity.\\
    Finally, whilst the performance of BERT and BioBERT is relatively similar, BioBERT converges faster across all tasks tested.
    
    \begin{table*}[!h]
    \begin{adjustbox}{width=.8\textwidth,center}
      \begin{tabular}{lllcccc}
        \hline
        & & & \multicolumn{4}{c}{\textbf{F1-score (average from 10-fold cross validation)}}\\
        \hline
        \textbf{Model} 
        & \textbf{Tokenizer} & \textbf{Embedding} 
        & \makecell{\textbf{MIMIC} \\ \textbf{Status}} 
        & \makecell{\textbf{MIMIC} \\ \textbf{Temporality} }
        & \makecell{\textbf{ShARe} \\ \textbf{Negation}} 
        & \makecell{\textbf{ShARe} \\ \textbf{Uncertainty}}\\
        \hline
        Bi-LSTM (custom) & SciSpacy & Wiki \textbar \ Word2Vec & 92.8\% & 97.3\% & 98.4\% & 96.7\%\\
        Bi-LSTM (custom) & SciSpacy & Wiki \textbar \ GloVe & 93.4\% & 97.2\% & 98.4\% & 97.2\%\\
        Bi-LSTM (custom) & SciSpacy & Wiki \textbar \ FastText & 93.6\% & 96.9\% & 98.6\% & 96.4\%\\
        Bi-LSTM (custom) & SciSpacy & MIMIC \textbar \ Word2Vec & \textbf{94.5}\% & \textbf{97.9}\% & \textbf{98.7}\% & \textbf{97.3}\%\\
        Bi-LSTM (custom) & SciSpacy & MIMIC \textbar \ GloVe & 93.9\% & \textbf{97.9}\% & \textbf{98.7}\% & 96.9\%\\
        Bi-LSTM (custom) & SciSpacy & MIMIC \textbar \ FastText & 93.7\% & 97.6\% & 98.5\% & 97.2\%\\
        BERT & WordPiece & BERTbase & 91.5\% & 97.3\% & 98.2\% & 93.6\%\\
        BioBERT & WordPiece & BioBERT & 93.4\% & 97.3\% & 98.5\% & 94.2\%\\
        SVM & SciSpacy & Wiki \textbar \ Word2Vec & 76.9\% & 94.8\% & 88.5\% & 85.9\%\\
        SVM & SciSpacy & Wiki \textbar \ GloVe & 78.6\% & 94.9\% & 88.8\% & 87.1\%\\
        SVM & SciSpacy & Wiki \textbar \ FastText & 78.1\% & 94.4\% & 88.7\% & 86.3\%\\
        SVM & SciSpacy & BoW & 82.7\% & 96.0\% & 90.2\% & 91.7\%\\
        SVM & SciSpacy & MIMIC \textbar \ Word2Vec & 80.6\% & 95.1\% & 89.8\% & 90.2\%\\
        SVM & SciSpacy & MIMIC \textbar \ GloVe & 79.1\% & 94.1\% & 89.4\% & 87.6\%\\
        SVM & SciSpacy & MIMIC \textbar \ FastText & 79.6\% & 93.7\% & 88.9\% & 88.0\%\\
        \hline
      \end{tabular}
    \end{adjustbox}
    \caption{Comparison of embeddings (steps A \& D)}
    \label{tab:embeddings}
    \end{table*}
    
    \begin{table*}[!h]
    \begin{adjustbox}{width=.8\textwidth,center}
      \begin{tabular}{lllcccc}
        \hline
        & & & \multicolumn{4}{c}{\textbf{F1-score (average from 10-fold cross validation)}}\\
        \hline
        \textbf{Model} 
        & \textbf{Tokenizer} & \textbf{Embedding} 
        & \makecell{\textbf{MIMIC} \\ \textbf{Status}} 
        & \makecell{\textbf{MIMIC} \\ \textbf{Temporality} }
        & \makecell{\textbf{ShARe} \\ \textbf{Negation}} 
        & \makecell{\textbf{ShARe} \\ \textbf{Uncertainty}}\\
        \hline
        Bi-LSTM & SciSpacy & MIMIC \textbar Word2Vec & 88.4\% & 97.1\% & 96.2\% & 94.1\%\\
        Bi-LSTM (custom) & SciSpacy & MIMIC \textbar Word2Vec & \textbf{94.5}\% & \textbf{97.9}\% & \textbf{98.7}\% & \textbf{97.3}\%\\
        BERT & WordPiece & BERTbase & 91.5\% & 97.3\% & 98.2\% & 93.6\%\\
        BioBERT & WordPiece & BioBERT & 93.4\% & 97.3\% & 98.5\% & 94.2\%\\
        ANN & SciSpacy & MIMIC \textbar Word2Vec & 80.9\% & 96.5\% & 88.6\% & 86.7\%\\
        CNN & SciSpacy & MIMIC \textbar Word2Vec & 84.6\% & 97.3\% & 92.0\% & 87.5\%\\
        RNN & SciSpacy & MIMIC \textbar Word2Vec & 77\% & 96.8\% & 94.0\% & 87.1\%\\
        RNN (custom) & SciSpacy & MIMIC \textbar Word2Vec & 89.5\% & 96.7\% & 97.9\% & 96.5\%\\
        SVM & SciSpacy & MIMIC \textbar Word2Vec & 80.6\% & 95.1\% & 89.8\% & 90.2\%\\
        ANN & SciSpacy & BoW & 79.8\% & 94.8\% & 89.3\% & 89.3\%\\
        SVM & SciSpacy & BoW & 82.7\% & 96\% & 90.2\% & 91.7\%\\
        \hline
      \end{tabular}
    \end{adjustbox}
    \caption{Comparison of classification algorithms (steps A \& D)}
    \label{tab:algos}
    \end{table*}

\subsection{Impact of text pre-processing (step B)}
    In addition to exploring various embeddings, we tested the impact of text pre-processing on classification task performance.
    In order to do so, we selected the best performing word embedding obtained in the previous step (Word2Vec trained on MIMIC-III, using SciSpacy tokenizer), and compared performances between all text cleaning variations (lowercasing, punctuation removal, stemming, lemmatization).\\
    For each variant investigated, the same pre-processing settings were applied to prepare the annotated corpus as well as to the entire MIMIC-III dataset, which was then used to re-train Word2Vec. This ensured the same vocabulary was used across the embedding and sequences to classify for each experiment.\\
    The results, summarized in Table~\ref{tab:preprocessing}, suggest that text pre-processing has a minor impact for all classification algorithms tested. Notably, stemming and lemmatization have a slightly negative impact on performance.
    
    \begin{table*}[!h]
    \centering
    \begin{adjustbox}{width=1\textwidth,center}
    \begin{tabular}{lllcccccc}
        \hline
        & & & \multicolumn{6}{c}{\textbf{F1-score (average from 10-fold cross validation)}}\\
        \hline
        \textbf{Task} & \textbf{Embedding} & \textbf{Text pre-processing} & \textbf{SVM} & \textbf{ANN} & \textbf{RNN} & \makecell{\textbf{RNN} (custom)} & \textbf{CNN} & \makecell{\textbf{Bi-LSTM} (custom)}  \\
        \hline
        MIMIC \textbar \ Status & MIMIC \textbar \ Word2Vec & Lowercase (L) & \textbf{80.6}\% & \textbf{80.9}\% & 77.0\% & \textbf{89.5}\% & 84.6\% & \textbf{94.5}\%\\
        MIMIC \textbar \ Status & MIMIC \textbar \ Word2Vec & L + punctuation removal (LP) & 80.1\% & 80.0\% & \textbf{80.2}\% & 86.1\% & \textbf{84.7}\% & 94.4\% \\
        MIMIC \textbar \ Status & MIMIC \textbar \ Word2Vec & LP + lemmatizing & \textbf{80.6}\% & 79.6\% & 78.0\% & 86.3\% & 83.8\% & 94.1\%\\
        MIMIC \textbar \ Status & MIMIC \textbar \ Word2Vec & LP + stemming & 80.4\% & 79.7\% & 79.4\% & 86.1\% & 84.1\% & 94.1\%\\
        \hline
    \end{tabular}
    \end{adjustbox}
    \caption{Comparison of text pre-processing methods (step B)}
    \label{tab:preprocessing}
    \end{table*}

\subsection{Impact of tokenizers (step C)}
    We tested the impact of tokenization on the performance of text classification tasks, focusing on SciSpacy and BBPE tokenizers, as they allow us to compare whole word versus subword unit methods.\\
    The results for the MIMIC \textbar \ Status task (and using Word2Vec trained on MIMIC-III)  are shown in Table~\ref{tab:tokenizers}, and indicate that the performances are roughly similar when using the BBPE tokenizer compared to SciSpacy.
    
    Furthermore we compared both approaches in terms of speed and vocabulary size.
    Tokenizing text took on average 2.5 times longer with Scispacy (250 seconds to tokenize 100,000 medical notes for SciSpacy versus 99 seconds for BBPE, excluding model loading time).
    For the models trained on MIMIC-III corpus, Scispacy comprised 474,145 words, and BBPE  29,452 subword units.
    
    \begin{table*}[!h]
    \centering
    \begin{adjustbox}{width=.9\textwidth,center}
    \begin{tabular}{lllcccccc}
        \hline
        & & & \multicolumn{6}{c}{\textbf{F1-score (average from 10-fold cross validation)}}\\
        \hline
        \textbf{Task} & \textbf{Embedding} & \textbf{Tokenizer} & \textbf{SVM} & \textbf{ANN} & \textbf{RNN}  & \makecell{\textbf{RNN} (custom)}& \textbf{CNN} & \makecell{\textbf{Bi-LSTM} (custom)}  \\
        \hline
        MIMIC \textbar \ Status & MIMIC \textbar \ Word2Vec & Sciscpacy & \textbf{80.6}\% & \textbf{80.9}\% & \textbf{77.0}\% & \textbf{89.5}\% & \textbf{84.6}\% & 94.5\%\\
        MIMIC \textbar \ Status & MIMIC \textbar \ Word2Vec & BBPE & 78.8\% & 80.5\% & 76.5\%  & 86.0\% & 84.3\% & \textbf{94.7}\% \\
        \hline
    \end{tabular}
    \end{adjustbox}
    \caption{Comparison of tokenizing methods (step C)}
    \label{tab:tokenizers}
    \end{table*}

\subsection{Embeddings analysis: word similarities comparison}
    Finally, in order to analyse the differences between embeddings trained on general and medical corpora, we compared the semantic information captured by Word2Vec (using SciSpacy tokenizer and without any preliminary text pre-processing).
    
    Table~\ref{tab:wordsim}  explores word similarities by showing the top ten similar words for medical (“cancer”) and non-medical (“concentration” and "attention") terms. \\
    Notably, it highlights the numerous misspellings, abbreviations and domain-specific meanings contained in the medical lexicon, suggesting that general corpora such as Wikipedia may not be appropriate when working on data from medical records (and by implication, for other specific domains).
    
    \begin{table*}[!h]
    \centering
    \begin{adjustbox}{width=.99\textwidth,center}
    \begin{tabular}{cc|cc|cc}
    \hline
        \multicolumn{2}{c}{\textbf{Term: “cancer”}} & \multicolumn{2}{c}{\textbf{Term: “concentration”}} & \multicolumn{2}{c}{\textbf{Term: “attention”}} \\
        \makecell{\textbf{Word2Vec Medical}}  & \makecell{\textbf{Word2Vec General}} & 
        \makecell{\textbf{Word2Vec Medical}}  & \makecell{\textbf{Word2Vec General}} & 
        \makecell{\textbf{Word2Vec Medical}}  & \makecell{\textbf{Word2Vec General}} 
        \\
        \hline
        ca (0.78) & prostate (0.85) & hmf (0.51) & concentrations (0.71) & paid (0.43) & attentions (0.65)\\
        carcinoma (0.78) & colorectal (0.82) & concentrations (0.49) & arbeitsdorf (0.67) & approximation (0.34) & notoriety (0.63)\\
        cancer- (0.75) & melanoma (0.8) & formula (0.47) & vulkanwerft (0.65) & followup (0.33) & attracted (0.63)\\
        caner (0.71) & pancreatic (0.8) & mct (0.47) & sophienwalde (0.64) & proximity (0.32) & criticism (0.63)\\
        adenocarcinoma (0.71) & leukemia (0.79) & polycose (0.47) & lagerbordell (0.64) & short-term (0.31) & publicity (0.57)\\
        ca- (0.64) & entity/breast\_cancer (0.79) & virtue (0.45) & sterntal (0.64) & mangagement (0.31) & praise (0.57)\\
        melanoma (0.64) & leukaemia (0.78) & corn (0.45) & dürrgoy (0.62) & atetntion (0.31) & aroused (0.56)\\
        cancer;dehydration (0.63) & tumour (0.77) & dosage (0.44) & straflager (0.61) & attnetion (0.3) & acclaim (0.55)\\
        cancer/sda (0.61) & cancers (0.76) & planimetry (0.44) & maidanek (0.61) & atention (0.3) & interest (0.55)\\
        rcc (0.61) & ovarian (0.75) & equation (0.44) & szebnie (0.61) & non-rotated (0.3) & admiration (0.55)\\
        \hline
    \end{tabular}
    \end{adjustbox}
    \caption{Comparison of word similarities between general and domain-specific embeddings}
    \label{tab:wordsim}
    \end{table*}

\FloatBarrier
\section{Discussion}
    This study compared the impact of various embedding and classification methods on four different text classification tasks.
    Notably we investigated the impact of pre-training embedding models on clinical corpora versus off-the-shelf models trained on general corpora.\vspace{5mm}
    
    The results suggest that using embeddings pre-trained for the specific task (clinical  corpora in our case) leads to better performance with any classification algorithm tested. However, pre-training such embeddings is not necessarily feasible due to either data or technical constraints. In this case our results highlight that using off-the-shelf embeddings trained on large general corpora such as Wikipedia still produce acceptable performance. In particular BERTbase outperformed most algorithms tested, even when these were combined with clinical embeddings.\\ 
    Additionally, BioBERT was not pre-trained on medical notes but on texts from a related domain (biomedical articles and abstracts as opposed to clinical records), and therefore excludes specificities inherent to the medical domain such as misspellings or technical jargon. Despite this, BioBERT's performance is only marginally below that of the best model (custom Bi-LSTM) combined with clinical embeddings.\vspace{5mm}
    
    The various experiments conducted on text pre-processing only lead to small variations in terms of performance, and even negatively impact the performance of several algorithms, for the text classification task and embedding model tested. Given the additional constraints required to perform this step (need to train embeddings on pre-processed texts and to clean input data) and the mixed results in performance, pre-processing does not appear to be essential.\vspace{5mm}
    
    Novel tokenization methods based on subword dictionaries, whilst not improving the performance, eliminate several shortcomings presented by SciSpacy and similar methods, notably its speed and vocabulary size.\\
    In light of these limitations and the very small difference in performance for the task tested, BBPE appears to be a suitable alternative to traditional tokenizers and allows to reduce significantly computational costs.\vspace{5mm}
    
    Finally, custom Bi-LSTM outperforms BioBERT when it simulates attention on the entity of interest. However, this configuration requires information on the entity mention span, and then to center each document on this span. For some datasets, such information may either be readily available, or can be obtained by performing an additional named-entity extraction step. Unfortunately, many text classification tasks do not usually have this information, or may not rely on the specific entities/keywords required (e.g. sentiment analysis tasks).
    When Bi-LSTM is not customized, then both BERT models (trained on general and specific domains) produce the best performance, and consequently should be preferred for texts not easily allowing such customization.\vspace{5mm}

\section{Conclusion}
    In this article we have explored the performance of various word representation approaches (comparing bag-of-words to traditional and contextual embeddings trained on both specific and general corpora, combined with various text pre-processing and tokenizing methods) as well as classification algorithms on four different text classification tasks, all based on publicly available datasets.\vspace{5mm}
    
    A detailed performance comparison on these four tasks highlighted the efficacy of contextual embeddings when compared to traditional methods when no customization is possible, whether these embeddings are trained on specific or general corpora.\\
    When combined with appropriate entity extraction tasks and specific domain embeddings, Bi-LSTM outperforms contextual embeddings.
    Across all classification algorithms, text pre-processing and tokenization approaches showed limited impact for the task and embedding tested, suggesting a rule of thumb to opt for the least time and resource intensive method.

\section*{Acknowledgments}
We thank the anonymous reviewers for their valuable suggestions.\\
This work was supported by Health Data Research UK, an initiative funded by UK Research and Innovation, Department of Health and Social Care (England) and the devolved administrations, and leading medical research charities.\\
AM is funded by Takeda California, Inc.\\
RD, RS, AR are part-funded by the National Institute for Health Research (NIHR) Biomedical Research Centre at South London and Maudsley NHS Foundation Trust and King's College London.\\
RD is also supported by The National Institute for Health Research University College London Hospitals Biomedical Research Centre, and by the BigData@Heart Consortium, funded by the Innovative Medicines Initiative-2 Joint Undertaking under grant agreement No. 116074. This Joint Undertaking receives support from the European Union’s Horizon 2020 research and innovation programme and EFPIA.\\
DB is funded by a UKRI Innovation Fellowship as part of Health Data Research UK MR/S00310X/1.\\
RB is funded in part by grant MR/R016372/1 for the King’s College London MRC Skills Development Fellowship programme funded by the UK Medical Research Council (MRC) and by grant IS-BRC-1215-20018 for the National Institute for Health Research (NIHR) Biomedical Research Centre at South London and Maudsley NHS Foundation Trust and King’s College London.\\

This paper represents independent research part funded by the National Institute for Health Research (NIHR) Biomedical Research Centre at South London and Maudsley NHS Foundation Trust and King’s College London. The views expressed are those of the authors and not necessarily those of the NHS, the NIHR or the Department of Health and Social Care. The funders had no role in study design, data collection and analysis, decision to publish, or preparation of the manuscript.\\

\bibliography{anthology,acl2020}

\begin{thebibliography}{31}
\expandafter\ifx\csname natexlab\endcsname\relax\def\natexlab#1{#1}\fi

\bibitem[{Bodenreider(2004)}]{bodenreider_unified_2004}
Olivier Bodenreider. 2004.
\newblock \href {https://doi.org/10.1093/nar/gkh061} {The {Unified} {Medical}
  {Language} {System} ({UMLS}): integrating biomedical terminology}.
\newblock \emph{Nucleic Acids Research}, 32(suppl\_1):D267--D270.

\bibitem[{Bojanowski et~al.(2017)Bojanowski, Grave, Joulin, and
  Mikolov}]{bojanowski_enriching_2017}
Piotr Bojanowski, Edouard Grave, Armand Joulin, and Tomas Mikolov. 2017.
\newblock \href {https://doi.org/10.1162/tacl_a_00051} {Enriching {Word}
  {Vectors} with {Subword} {Information}}.
\newblock \emph{Transactions of the Association for Computational Linguistics},
  5:135--146.

\bibitem[{Chapman et~al.(2001)Chapman, Bridewell, Hanbury, Cooper, and
  Buchanan}]{chapman_simple_2001}
Wendy~W. Chapman, Will Bridewell, Paul Hanbury, Gregory~F. Cooper, and Bruce~G.
  Buchanan. 2001.
\newblock \href {https://doi.org/10.1006/jbin.2001.1029} {A {Simple}
  {Algorithm} for {Identifying} {Negated} {Findings} and {Diseases} in
  {Discharge} {Summaries}}.
\newblock \emph{Journal of Biomedical Informatics}, 34(5):301--310.

\bibitem[{Cortes and Vapnik(1995)}]{cortes_support-vector_1995}
Corinna Cortes and Vladimir Vapnik. 1995.
\newblock \href {https://doi.org/10.1007/BF00994018} {Support-vector networks}.
\newblock \emph{Machine Learning}, 20(3):273--297.

\bibitem[{Dai(2019)}]{dai_family_2019}
Hong-Jie Dai. 2019.
\newblock \href {https://doi.org/10.1186/s12911-019-0996-4} {Family member
  information extraction via neural sequence labeling models with different tag
  schemes}.
\newblock \emph{BMC Medical Informatics and Decision Making}, 19(10):257.

\bibitem[{Devlin et~al.(2019)Devlin, Chang, Lee, and
  Toutanova}]{devlin_bert_2019}
Jacob Devlin, Ming-Wei Chang, Kenton Lee, and Kristina Toutanova. 2019.
\newblock \href {http://arxiv.org/abs/1810.04805} {{BERT}: {Pre}-training of
  {Deep} {Bidirectional} {Transformers} for {Language} {Understanding}}.
\newblock \emph{arXiv:1810.04805 [cs]}.
\newblock ArXiv: 1810.04805.

\bibitem[{Elhadad et~al.(2013)Elhadad, Chapman, O’Gorman, Palmer, and
  Savova}]{elhadad_share_2013}
N~Elhadad, W.~W Chapman, T~O’Gorman, M~Palmer, and G~Savova. 2013.
\newblock The {ShARe} schema for the syntactic and semantic annotation of
  clinical texts.

\bibitem[{Gage(1994)}]{gage_new_1994}
Philip Gage. 1994.
\newblock A new algorithm for data compression.

\bibitem[{Harkema et~al.(2009)Harkema, Dowling, Thornblade, and
  Chapman}]{harkema_context_2009}
Henk Harkema, John~N. Dowling, Tyler Thornblade, and Wendy~W. Chapman. 2009.
\newblock \href {https://doi.org/10.1016/j.jbi.2009.05.002} {{ConText}: an
  algorithm for determining negation, experiencer, and temporal status from
  clinical reports}.
\newblock \emph{Journal of Biomedical Informatics}, 42(5):839--851.

\bibitem[{Johnson et~al.(2016)Johnson, Pollard, Shen, Lehman, Feng, Ghassemi,
  Moody, Szolovits, Celi, and Mark}]{johnson_mimic-iii_2016}
Alistair E.~W. Johnson, Tom~J. Pollard, Lu~Shen, Li-wei~H. Lehman, Mengling
  Feng, Mohammad Ghassemi, Benjamin Moody, Peter Szolovits, Leo~Anthony Celi,
  and Roger~G. Mark. 2016.
\newblock \href {https://doi.org/10.1038/sdata.2016.35} {{MIMIC}-{III}, a
  freely accessible critical care database}.
\newblock \emph{Scientific Data}, 3(1):1--9.

\bibitem[{Kenter et~al.(2016)Kenter, Borisov, and
  de~Rijke}]{kenter_siamese_2016}
Tom Kenter, Alexey Borisov, and Maarten de~Rijke. 2016.
\newblock \href {https://doi.org/10.18653/v1/P16-1089} {Siamese {CBOW}:
  {Optimizing} {Word} {Embeddings} for {Sentence} {Representations}}.
\newblock In \emph{Proceedings of the 54th {Annual} {Meeting} of the
  {Association} for {Computational} {Linguistics} ({Volume} 1: {Long}
  {Papers})}, pages 941--951, Berlin, Germany. Association for Computational
  Linguistics.

\bibitem[{Koleck et~al.(2019)Koleck, Dreisbach, Bourne, and
  Bakken}]{koleck_natural_2019}
T.A. Koleck, C.~Dreisbach, P.E. Bourne, and S.~Bakken. 2019.
\newblock \href {https://doi.org/10.1093/jamia/ocy173} {Natural language
  processing of symptoms documented in free-text narratives of electronic
  health records: {A} systematic review}.
\newblock \emph{Journal of the American Medical Informatics Association},
  26(4):364--379.

\bibitem[{Kraljevic et~al.(2019)Kraljevic, Bean, Mascio, Roguski, Folarin,
  Roberts, Bendayan, and Dobson}]{kraljevic_medcat_2019}
Zeljko Kraljevic, Daniel Bean, Aurelie Mascio, Lukasz Roguski, Amos Folarin,
  Angus Roberts, Rebecca Bendayan, and Richard Dobson. 2019.
\newblock \href {http://arxiv.org/abs/1912.10166} {{MedCAT} -- {Medical}
  {Concept} {Annotation} {Tool}}.
\newblock \emph{arXiv:1912.10166 [cs, stat]}.
\newblock ArXiv: 1912.10166.

\bibitem[{Lee et~al.(2019)Lee, Yoon, Kim, Kim, Kim, So, and
  Kang}]{lee_biobert_2019}
Jinhyuk Lee, Wonjin Yoon, Sungdong Kim, Donghyeon Kim, Sunkyu Kim, Chan~Ho So,
  and Jaewoo Kang. 2019.
\newblock \href {https://doi.org/10.1093/bioinformatics/btz682} {{BioBERT}: a
  pre-trained biomedical language representation model for biomedical text
  mining}.
\newblock \emph{Bioinformatics}, page btz682.
\newblock ArXiv: 1901.08746.

\bibitem[{Meystre et~al.(2019)Meystre, Heider, Kim, Aruch, and
  Britten}]{meystre_automatic_2019}
Stéphane~M. Meystre, Paul~M. Heider, Youngjun Kim, Daniel~B. Aruch, and
  Carolyn~D. Britten. 2019.
\newblock \href {https://doi.org/10.1016/j.ijmedinf.2019.05.018} {Automatic
  trial eligibility surveillance based on unstructured clinical data}.
\newblock \emph{International Journal of Medical Informatics}, 129:13--19.

\bibitem[{Mikolov et~al.(2013)Mikolov, Chen, Corrado, and
  Dean}]{mikolov_efficient_2013}
Tomas Mikolov, Kai Chen, Greg Corrado, and Jeffrey Dean. 2013.
\newblock \href {http://arxiv.org/abs/1301.3781} {Efficient {Estimation} of
  {Word} {Representations} in {Vector} {Space}}.
\newblock \emph{arXiv:1301.3781 [cs]}.
\newblock ArXiv: 1301.3781.

\bibitem[{Mowery et~al.(2014)Mowery, Velupillai, South, Christensen, Martinez,
  Kelly, Goeuriot, Elhadad, Savova, and Chapman}]{mowery_task_2014}
Danielle~L Mowery, Sumithra Velupillai, Brett~R South, Lee Christensen, David
  Martinez, Liadh Kelly, Lorraine Goeuriot, Noemie Elhadad, Guergana Savova,
  and Wendy~W Chapman. 2014.
\newblock Task 2: {ShARe}/{CLEF} {eHealth} {Evaluation} {Lab} 2014.
\newblock page~12.

\bibitem[{Mujtaba et~al.(2019)Mujtaba, Shuib, Idris, Hoo, Raj, Khowaja, Shaikh,
  and Nweke}]{mujtaba_clinical_2019}
Ghulam Mujtaba, Liyana Shuib, Norisma Idris, Wai~Lam Hoo, Ram~Gopal Raj, Kamran
  Khowaja, Khairunisa Shaikh, and Henry~Friday Nweke. 2019.
\newblock \href {https://doi.org/10.1016/j.eswa.2018.09.034} {Clinical text
  classification research trends: {Systematic} literature review and open
  issues}.
\newblock \emph{Expert Systems with Applications}, 116:494--520.

\bibitem[{Neumann et~al.(2019)Neumann, King, Beltagy, and
  Ammar}]{neumann_scispacy_2019}
Mark Neumann, Daniel King, Iz~Beltagy, and Waleed Ammar. 2019.
\newblock \href {https://doi.org/10.18653/v1/W19-5034} {{ScispaCy}: {Fast} and
  {Robust} {Models} for {Biomedical} {Natural} {Language} {Processing}}.
\newblock \emph{Proceedings of the 18th BioNLP Workshop and Shared Task}, pages
  319--327.
\newblock ArXiv: 1902.07669.

\bibitem[{Pennington et~al.(2014)Pennington, Socher, and
  Manning}]{pennington_glove:_2014}
Jeffrey Pennington, Richard Socher, and Christopher Manning. 2014.
\newblock \href {https://doi.org/10.3115/v1/D14-1162} {Glove: {Global}
  {Vectors} for {Word} {Representation}}.
\newblock In \emph{Proceedings of the 2014 {Conference} on {Empirical}
  {Methods} in {Natural} {Language} {Processing} ({EMNLP})}, pages 1532--1543,
  Doha, Qatar. Association for Computational Linguistics.

\bibitem[{Purushotham et~al.(2018)Purushotham, Meng, Che, and
  Liu}]{purushotham_benchmarking_2018}
Sanjay Purushotham, Chuizheng Meng, Zhengping Che, and Yan Liu. 2018.
\newblock \href {https://doi.org/10.1016/j.jbi.2018.04.007} {Benchmarking deep
  learning models on large healthcare datasets}.
\newblock \emph{Journal of Biomedical Informatics}, 83:112--134.

\bibitem[{Saeed et~al.(2002)Saeed, Lieu, Raber, and Mark}]{saeed_mimic_2002}
M.~Saeed, C.~Lieu, G.~Raber, and R.~G. Mark. 2002.
\newblock {MIMIC} {II}: a massive temporal {ICU} patient database to support
  research in intelligent patient monitoring.
\newblock \emph{Computers in Cardiology}, 29:641--644.

\bibitem[{Sennrich et~al.(2016)Sennrich, Haddow, and
  Birch}]{sennrich_neural_2016}
Rico Sennrich, Barry Haddow, and Alexandra Birch. 2016.
\newblock \href {http://arxiv.org/abs/1508.07909} {Neural {Machine}
  {Translation} of {Rare} {Words} with {Subword} {Units}}.
\newblock \emph{arXiv:1508.07909 [cs]}.
\newblock ArXiv: 1508.07909.

\bibitem[{Si et~al.(2019)Si, Wang, Xu, and Roberts}]{si_enhancing_2019}
Yuqi Si, Jingqi Wang, Hua Xu, and Kirk Roberts. 2019.
\newblock \href {https://doi.org/10.1093/jamia/ocz096} {Enhancing clinical
  concept extraction with contextual embeddings}.
\newblock \emph{Journal of the American Medical Informatics Association},
  26(11):1297--1304.

\bibitem[{Uzuner et~al.(2008)Uzuner, Goldstein, Luo, and
  Kohane}]{uzuner_identifying_2008}
Ozlem Uzuner, Ira Goldstein, Yuan Luo, and Isaac Kohane. 2008.
\newblock \href {https://doi.org/10.1197/jamia.M2408} {Identifying patient
  smoking status from medical discharge records}.
\newblock \emph{Journal of the American Medical Informatics Association:
  JAMIA}, 15(1):14--24.

\bibitem[{Uzuner(2009)}]{uzuner_recognizing_2009}
Özlem Uzuner. 2009.
\newblock \href {https://doi.org/10.1197/jamia.M3115} {Recognizing {Obesity}
  and {Comorbidities} in {Sparse} {Data}}.
\newblock \emph{Journal of the American Medical Informatics Association :
  JAMIA}, 16(4):561--570.

\bibitem[{Wang et~al.(2019)Wang, Cho, and Gu}]{wang_neural_2019}
Changhan Wang, Kyunghyun Cho, and Jiatao Gu. 2019.
\newblock \href {http://arxiv.org/abs/1909.03341} {Neural {Machine}
  {Translation} with {Byte}-{Level} {Subwords}}.
\newblock \emph{arXiv:1909.03341 [cs]}.
\newblock ArXiv: 1909.03341.

\bibitem[{Wang et~al.(2018)Wang, Wang, Rastegar-Mojarad, Moon, Shen, Afzal,
  Liu, Zeng, Mehrabi, Sohn, and Liu}]{wang_clinical_2018}
Y.~Wang, L.~Wang, M.~Rastegar-Mojarad, S.~Moon, F.~Shen, N.~Afzal, S.~Liu,
  Y.~Zeng, S.~Mehrabi, S.~Sohn, and H.~Liu. 2018.
\newblock \href {https://doi.org/10.1016/j.jbi.2017.11.011} {Clinical
  information extraction applications: {A} literature review}.
\newblock \emph{Journal of Biomedical Informatics}, 77:34--49.

\bibitem[{Wolf et~al.(2020)Wolf, Debut, Sanh, Chaumond, Delangue, Moi, Cistac,
  Rault, Louf, Funtowicz, and Brew}]{wolf_huggingfaces_2020}
Thomas Wolf, Lysandre Debut, Victor Sanh, Julien Chaumond, Clement Delangue,
  Anthony Moi, Pierric Cistac, Tim Rault, Rémi Louf, Morgan Funtowicz, and
  Jamie Brew. 2020.
\newblock \href {http://arxiv.org/abs/1910.03771} {{HuggingFace}'s
  {Transformers}: {State}-of-the-art {Natural} {Language} {Processing}}.
\newblock \emph{arXiv:1910.03771 [cs]}.
\newblock ArXiv: 1910.03771.

\bibitem[{Yamada et~al.(2018)Yamada, Asai, Shindo, Takeda, and
  Takefuji}]{yamada_wikipedia2vec_2018}
Ikuya Yamada, Akari Asai, Hiroyuki Shindo, Hideaki Takeda, and Yoshiyasu
  Takefuji. 2018.
\newblock \href {http://arxiv.org/abs/1812.06280} {{Wikipedia2Vec}: {An}
  {Optimized} {Tool} for {Learning} {Embeddings} of {Words} and {Entities} from
  {Wikipedia}}.
\newblock \emph{arXiv:1812.06280 [cs]}.
\newblock ArXiv: 1812.06280.

\bibitem[{Yao et~al.(2019)Yao, Mao, and Luo}]{yao_clinical_2019}
Liang Yao, Chengsheng Mao, and Yuan Luo. 2019.
\newblock \href {https://doi.org/10.1186/s12911-019-0781-4} {Clinical text
  classification with rule-based features and knowledge-guided convolutional
  neural networks}.
\newblock \emph{BMC Medical Informatics and Decision Making}, 19(3):71.

\end{thebibliography}
\bibliographystyle{acl_natbib}

\appendix

\section{Appendices}
\label{sec:appendix}
\subsection{\small set comparison of word representation, text pre-processing, tokenization and classification methods across tasks}.
\label{sec:appendix1}

\end{document}